\documentclass[11pt,a4paper]{article}
\usepackage[hyperref]{acl2021}
\usepackage{times}
\usepackage{latexsym}
\usepackage{graphicx} 
\usepackage{amssymb}
\usepackage{amsmath}
\usepackage{float}
\usepackage{subfigure}
\usepackage[capitalise]{cleveref}

\usepackage{microtype}

\aclfinalcopy 
\usepackage{color}

\title{MagicPai at SemEval-2021 Task 7: Method for Detecting and Rating Humor Based on Multi-Task Adversarial Training}

\author{Jian Ma, Shuyi Xie, Haiqin Yang$^{\S}$, Lianxin Jiang,\\ {\bf Mengyuan Zhou}, {\bf Xiaoyi Ruan}, {\bf Yang Mo} \\
         Ping An Life Insurance, Ltd. \\
  Shenzhen, Guangdong province, China \\
         {\small\{MAJIAN446,XIESHUYI542,JIANGLIANXIN769,RUANXIAOYI687,MOYANG853\}@pingan.com.cn}
         \\$^{\S}$ {\small the corresponding author, email: {hqyang@ieee.org}}
         }
\date{}

\begin{document}
\maketitle
\begin{abstract}
This paper describes MagicPai’s system for SemEval 2021 Task 7, HaHackathon: Detecting and Rating Humor and Offense. This task aims to detect whether the text is humorous and how humorous it is.  There are four subtasks in the competition.  In this paper, we mainly present our solution, a multi-task learning model based on adversarial examples, for task 1a and 1b.  More specifically, we first vectorize the cleaned dataset and add the perturbation to obtain more robust embedding representations.  We then correct the loss via the confidence level.  Finally, we perform interactive joint learning on multiple tasks to capture the relationship between whether the text is humorous and how humorous it is. The final result shows the effectiveness of our system.
\end{abstract} 

\section{Introduction}

Humor is the tendency of experiences to provoke laughter and provide amusement.  Regardless of gender, age or cultural background, it is a special way of language expression to provide an active atmosphere or resolve embarrassment in life while being an important medium for maintaining mental health~\cite{19}.  Recently, with the rapid development of artificial intelligence, it becomes one of the most hot research topics in natural language processing to recognize humor~\cite{20}.  The task of humor recognition consists of two subtasks: whether the text contains humorous and what level of the humor it is.  Early humor recognition methods tackle this task mainly by designing heuristic humor-specific features on classification models~\cite{18} and have proved that this automatic way can attain satisfactory performance.  Nowadays, researchers try to resolve this task by statistical machine learning or deep learning technologies.

The SemEval 2021 Task 7, HaHackathon: Detecting and Rating Humor and Offense, consists of four subtasks: Subtask 1 simulates the previous humor detection task, in which all scores are averaged to provide an average classification score.  Subtask 1a is a binary classification task to detect whether the text is humorous.  Subtask 1b is a regression task to predict how humorous it is for ordinary users in a value range from 0 to 5.  Subtask 1c is also a binary classification task to predict whether the humor grade causes controversy if the text is classified as humorous.  Subtask 2 aims to predict how offensive text for an ordinary user is in an integral value range from 0 and 5.

Due to the highly subjective nature of humor detection, the data is labeled by people with different profile in gender, age group, political position, income level, social status, etc. The tasks are extremely challenging because they lack a unified standard to define humor.

To tackle the tasks, we first preprocess the text, including stemming, acronym reduction, etc.  We then apply the pre-trained language model to get the representation of each subword in the text as the model input.  Meanwhile, we add a perturbation to the embedding layer and design an optimization goal that maximizes the perturbation of the loss function.  After that, we perform interactive multi-task learning on judging whether humor exists and predicting how humorous it is.  That is, based on maximizing the likelihood estimation under the Gaussian distribution with the same variance, we construct a multi-task loss function and automatically select different loss weights in the learning to improve the accuracy of each task.

\section{Related Work}

The early stages of humor recognition are based on statistical machine learning methods.  For example, \citet{1} try to learn statistical patterns of text in N-grams and provide a heuristic focus for a location of where wordplay may or may not occur. \citet{2} show that automatic classification techniques can be effectively deploy to distinguish between humorous and non-humorous texts and obtain significant improvement over the Apriori algorithm, a well-known baseline.  In addition, three human-centric features are designed for recognizing humor in the curated one-liner dataset. \citet{3} apply SVM models for humor recognition as a binary classification task and prove that the technique of metaphorical mapping can be generalized to identify other types of double entendre and other forms of humor. \citet{4} present several modifications of the original recurrent neural network language model to solve the humor recognition task. \citet{5} collect a crowdsourced corpus for humor classification from Spanish tweets and conduct extensive experiments to compare various machine learning models, such as  Support Vector Machine (SVM), a Multinomial version of Na\"ive Bayes (MNB), Decision Trees (DT), k Nearest Neighbors (kNN), and a
Gaussian version of Naïve Bayes (GNB).  \citet{6} observe that bigram language models performed slightly better than trigram models and there is some evidence that neural network models can outperform standard back-off N-gram models. \citet{7} extend the techniques of automatic humor recognition to different types of humor as well as different languages in both English and Chinese and proposed a deep learning CNN architecture with high way networks that can learn to distinguish between humorous and nonhumorous texts based on a large scale of balanced positive and negative dataset.

With the rapid development of deep learning technology, various pre-training models have made great progress in the field of natural language processing~\cite{conf/ijcnn/EDM21,conf/ijcnn/RefBERT21,conf/ijcai/PHED21}. \citet{8} propose to model sentiment association between elementary discourse units and compare various CNN methods of humor recognition. \citet{9} employ a Transformer architecture for its advantages in learning from sentence context and demonstrate the effectiveness of this approach and show results that are comparable to human performance. \citet{10} propose a new algorithm Enhancement Inference BERT (EI-BERT) that performs well in sentence classification. \citet{11} propose an internal and external attention neural network (IEANN) 
Attention mechanism~\cite{11,DBLP:conf/naacl/JiaoYKL19} has been applied and show good model performance.  The existing work can be borrowed or inspired our proposal in this paper.

\begin{figure*}[ht]
\centering
\subfigure[Perturbation embeddings]{\includegraphics[width=0.55\textwidth]{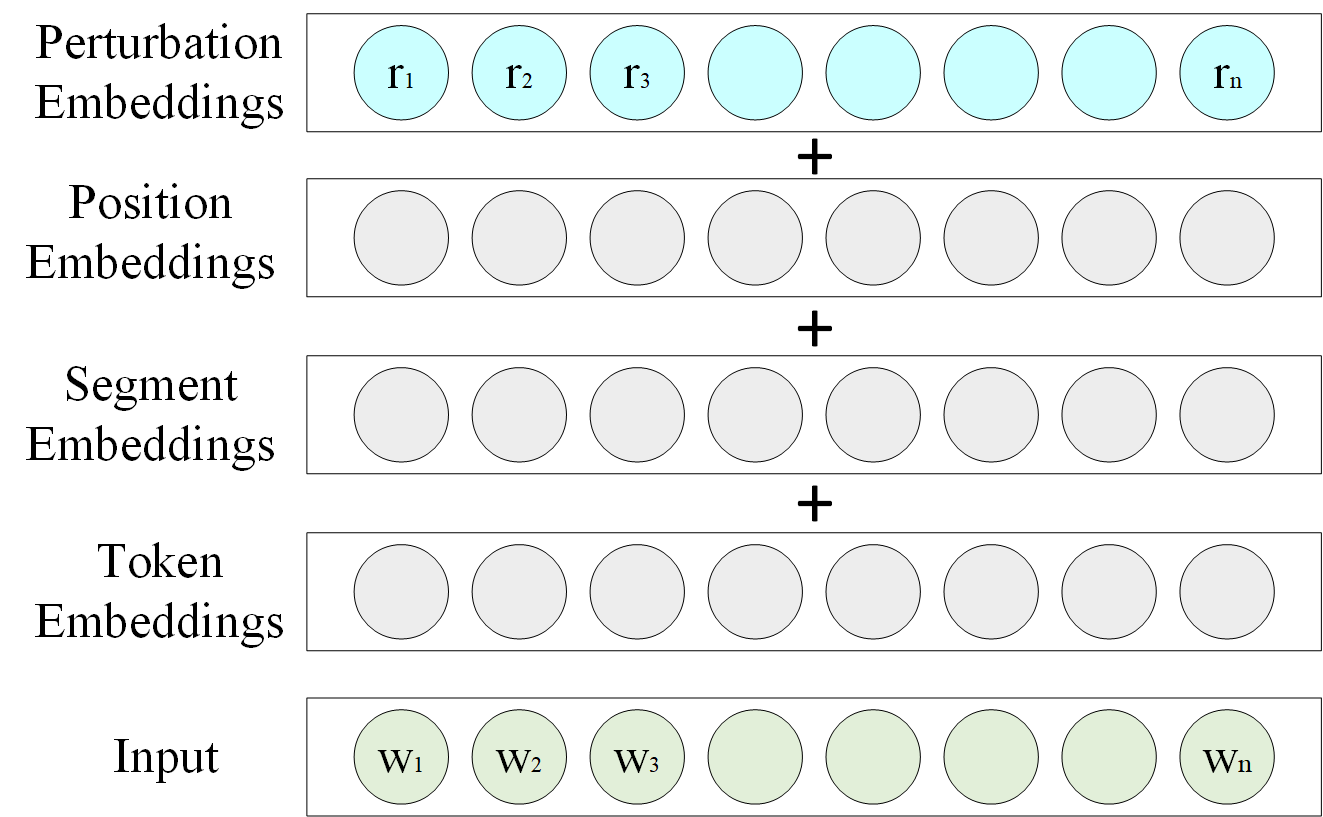}\label{fig:pe}}
\subfigure[Interactive multi-task learning]{\includegraphics[width=0.4\textwidth]{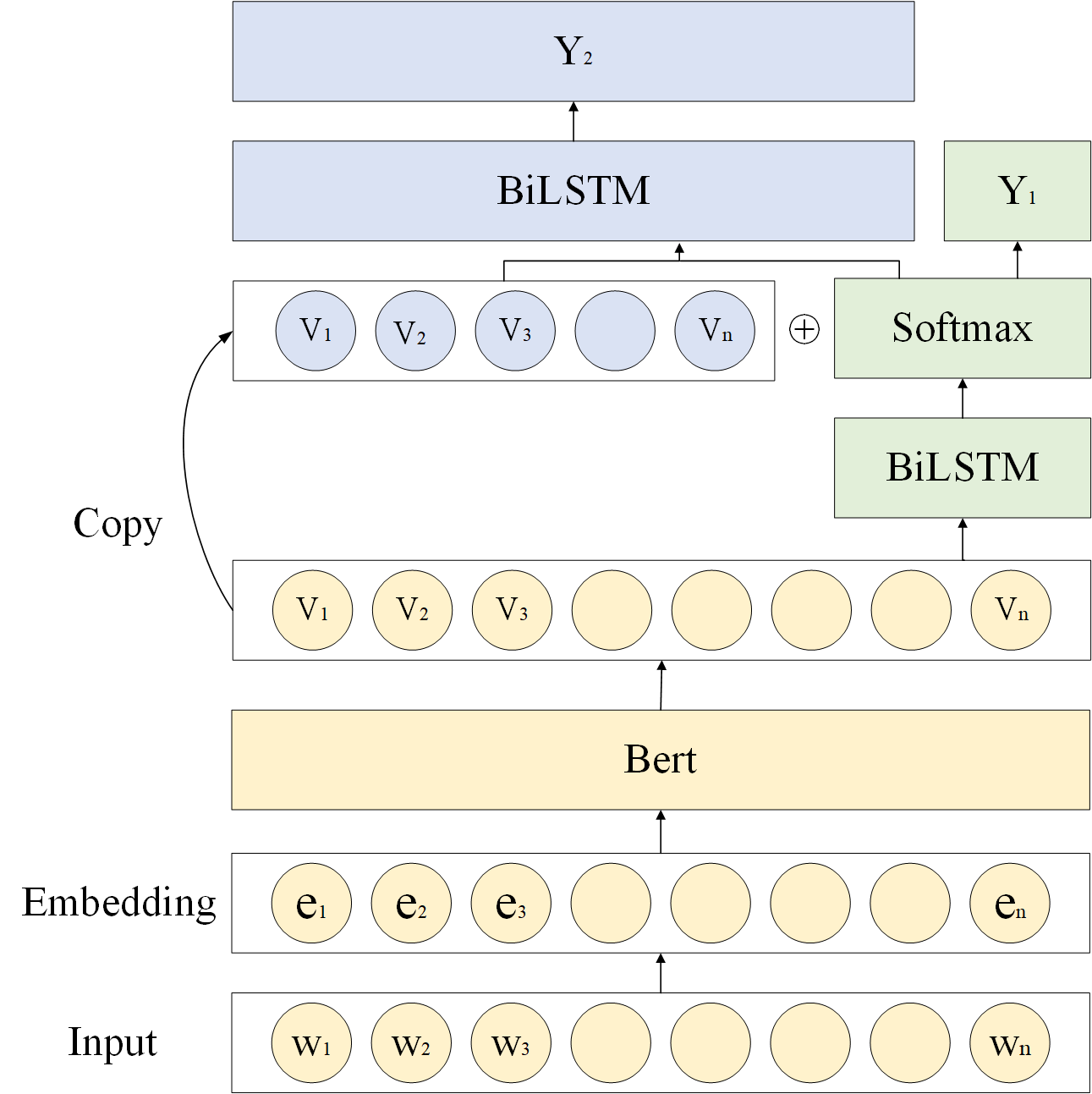}\label{fig:imtl}}
\caption{The main implementation of our proposed system.}
\end{figure*}
\section{Overview}
In the following, we present the implementation of our system for the competition.
\subsection{Virtual Adversarial Training Based on Loss Correction}

{
Recently, adversarial examples~\cite{12} have been generated to increase the robustness of training deep learning models~\cite{pan2019recent,lei2020conversational}.  This work is motivated by the significant discontinuities between the input-output mappings of deep neural networks.   When an imperceptible perturbation is added to the input, it may make the original normal network misclassify the result.} The characteristics of these perturbations are not random artifacts of learning generated by the network during the learning process, because the same perturbation will cause different networks trained on different data sets to produce the same classification errors.  Adversarial examples are samples that significantly improve the loss of the model by adding small perturbations to the input samples.

{
The adversarial training~\cite{13} is a training process that can effectively identify the original sample and the adversarial sample model.  Usually, the adversarial training requires labeled samples to provide supervision loss because the perturbation is designed to increase the model loss function.  Virtual adversarial training~\cite{14} extends the adversarial training to semi-supervised mode by adding regularization to the model so that the output distribution of a sample is the same as the output distribution after perturbation while attaining good performance in both supervised and unsupervised tasks.} When the training sample is mixed with noise, it is easy to overfit the model and learn wrong information. Therefore, it is necessary to interfere to control the influence of noise.

{
\Cref{fig:pe} illustrates the perturbation in our implementation.  For a word with a sequence length of ${n}$, we let ${w_i}$ denote the $i$-th subword, where $i=1, \ldots, n$. The representation of $w_i$ is then computed by the sum of token embedding, segmentation embedding, position embedding, and perturbation embedding, an additional embedding.  This makes it slightly different from the existing pre-trained language models, e.g., BERT. 
}

{
The virtual adversarial training can be unified by the following objective:
\begin{align} \nonumber
\min_{\theta}\mathbb{E}_{(x,y){\sim}D} [{\alpha} \max_{\beta} &L(f(x+{\beta};{\theta}),f(x;\theta))\\\label{eq:VAT} &+L(f(x;\theta),y)],
\end{align}
where $D$ is a training dataset consisting of input-output pairs $(x,y)$, $\alpha$ is a hyperparameter to control the trade-off between the standard error and the robust error. $\beta$ is the  adversarial perturbation, $y$ represents the true label, $\theta$ is the model parameter, $L$ is the loss function.  $x+\beta$ quantifies the perturbation $\beta$ injecting into $x$.  The goal of $\beta$ is to maximize the difference between the two decision function values, $f(x+{\beta};{\theta})$ and $f(x;{\theta})$, i.e., to make the prediction of the existing model as incorrect as possible.  To make ${\beta}$ meet a certain constraint, a conventional setting is to let ${\|\beta\|\leq\epsilon}$, where $\epsilon$ is a constant.  After constructing an adversarial sample $x+{\beta}$ for each sample, Eq.~(\ref{eq:VAT}) tries to seek the model parameter $\theta$ by minimizing the prediction loss. 

Since the training samples are mixed with noise, it is easy for the model to overfit and learn wrong information~\cite{17}, interference is adopted to control the influence of noise.  The loss function is defined as follows:
\begin{align} \label{eq:interference}
\mathcal{L}=-\sum_{i=1}^{N}((1-w_i)y_i+w_i\tilde{y}_i)\log(l_i)
\end{align}
where $y_i$ is the true label, $\tilde{y}_i$ is the predicted label, and $l_i$ is the predicted probability distribution. $w_i$ is a hyperparameter to control the trade-off between true label and predicted label.  By minimizing the loss defined in Eq.~(\ref{eq:interference}), we can reduce the attention to noise points by adding the model's own predictions to the true labels and the prediction to the noise point. 
}

\subsection{Interactive Multi-task Training}
According to the description of the first two tasks, task 1a is a binary classification task to predict if the text would be considered humorous for an average user while task 1b is a regression task to determine how humorous it is for an average user when the text is classed as humorous, where the values vary between 0 and 5.  In order to capture the relationship between whether text is humorous and how humorous it is, we designed the network structure shown in \cref{fig:imtl}.  {The input, as illustrated in \cref{fig:pe}, is the sum of the token embedding, position embedding, segment embedding, and perturbation embedding.  The sum of four embeddings is sent to a pre=trained language model (PLM) to yield an input for a BiLSTM model.  After that, a Softmax  layer is placed to recognize whether the text is humor.  Meanwhile, the output of the PLM and the output of the Softmax layer are concatenated together and sent to another BiSLTM model to predict how humorous it is.  In \cref{fig:imtl}, the notation $\oplus$ represents the concatenation operation.  Because two tasks have different noise patterns, learning two tasks simultaneously can make features interact in the tasks.  For task 1a, it is easy to learn some important features while for task 1b, it is difficult to extract them.  The reason may come from the following facts: the interaction between task 1b and the features may be too complicated, or some other features may hinder the learning procedure~\cite{15}.  Hence, by deploying interactive multi-task learning, we can get a more generalized representation.
}

{
Since different loss functions have different scales, loss functions with a larger scale will significantly dominate the loss functions with a smaller scale~\cite{liang2020pirhdy,zhang2020generalized}. Therefore, a weighted summation of the loss function is required to make balance on the loss functions.  Motivating by \cite{16} that modeling is based on task-dependent and homoscedastic aleatoric uncertainty, i.e., for a certain sample, the model not only predicting its label but also estimating the task-dependent homoscedastic uncertainty, we present a multi-task loss function derived by maximizing the Gaussian likelihood of the same variance uncertainty.  Suppose the input is $X$, the parameter matrix $W$ is the model parameter for the output, $f^W(x)$.
For the classification in task 1a, the Softmax likelihood can be defined by:
\begin{align} 
p(y_1|f^W(x))=Softmax(f^W(x),\sigma_1),
\end{align}
where $\sigma_1$ is the observed noise scalar for the classification model.

For the regression task in task 1b, we can define its probability as the Gaussian likelihood by:
\begin{align} 
p(y_2|f^W(x))=G(f^W(x),\sigma_2),
\end{align}
where $\sigma_2$ is the observed noise scalar for the regression model.  

Here, to learn the models in the multi-task mode, we define the multivariate probability by
\begin{align}\label{eq:MTL_prob.}
&p(y1,y2|f^W(x))\\\nonumber
=&p(y_1|f^W(x))\cdot p(y_2|f^W(x))\\\nonumber
=&Softmax(f^W(x),\sigma_1) \cdot G(f^W(x),\sigma_2). 
\end{align}
Maximizing the probability defined in Eq.~(\ref{eq:MTL_prob.}) is equivalent to minimizing the following objective:
\begin{align}
&L(W,\sigma_1,\sigma_2) \\\nonumber=& -\log p(y1,y2|f^W(x))\\\nonumber
=&-\log Softmax(f^W(x),\sigma_1) \cdot G(f^W(x),\sigma_2)\\\nonumber
\propto & \frac{1}{\sigma_1^2}L_1(W)+\frac{1}{2\sigma_2^2}L_2(W)+\log \sigma_1+\log \sigma_2
\end{align} 
where $L_1=-\log Softmax(f^W(x),y_1)$ defines the cross entropy loss between the prediction and $y_1$.  $L_2=\|y_2-f^W(x)\|^2$ defines the Euclidean loss between the prediction and $y_2$.  By minimizing the above objective, we can learn the parameters of $W$, $\sigma_1$, and $\sigma_2$ accordingly.   
}

\begin{table}[!htb]
\centering
\begin{tabular}{@{~}c@{~~~}c@{~~}c@{~~}c@{~~}c@{~~~}c@{~~}c@{~~~}c@{~~}c@{~~}c@{~}}
\hline  & \multicolumn{3}{c}{\textbf{Train}} & \multicolumn{3}{c}{\textbf{Dev}} & \multicolumn{3}{c}{\textbf{Test}} \\ \hline
& No. & R. & L. & No. & R. & L. & No. & R. & L. \\\hline
1a & 8000 & 7:3 & 20 & 1000 & 5:3 & 19 & 1000 & * & 23\\
1b & 4935 &  *  & 19 & 1000 &  *  & 19 & 1000 & * & 23\\
1c & 4935 & 1:1 & 19 & 1000 & 1:1 & 19 & 1000 & * & 23\\
2 & 8000 &  *  & 20 & 1000 &  *  & 19 & 1000 & * & 23\\
\hline
\end{tabular}
\caption{\label{tab:stat}Data Statistics.  R.: The ratio of positive and negative samples.  L.: the average length.  * indicates the data is unavailable. }
\end{table}

\begin{table}[!htb]
\centering
\begin{tabular}{rcccc}
\hline  & \textbf{BERT} & \textbf{RoBERTa}& \textbf{XLNet}& \textbf{ERNIE} \\ \hline
lr& 2e-5& 5e-6& 5e-6& 3e-5\\
nte & 5& 10& 10& 15\\
bs & 64& 32& 32& 32\\
msl & 128& 100& 80& 80\\
wp & 0.05& 0.1& 0.05& 0.05\\
\hline
\end{tabular}
\caption{\label{tab:para}Parameters for different pre-trained language models.  lr: learning rate.  nte: no. of training epochs.  bs: batch size.  msl: max. sequence length. wp: warmup proportion. }
\end{table}
\section{Experiments}
\label{sec:length}
In the following, we present the data, experimental setup and analyze the results.

\subsection{Data and Experimental Setup}
{
The data is collected from the official release in~\cite{meaney2021hahackathon}.  We preprocess the data by spelling correction, stemming, handling special symbols, and converting all letters to lowercase, etc.  Finally, we obtain the data and report the statistics in Table~\ref{tab:stat}.

In the experiment, we choose the large version of four popular pre-training language models, i.e., BERT, XLNet, RoBERTa, and ERNIE.  The hyperparameters of each model are tuned based on our experience and shown in Table~\ref{tab:para}.  To train a good classifier, we deliver the following procedure: 1) conducting five-fold cross-validation on the training set and obtaining 20 models; 2) applying the 20 models to get the pseudo-labels of the data in the test set and extracting the data with high confidence, i.e., the predicted label score greater than 0.95 or smaller than 0.05, as new training data; 3) the pseudo label data from the test set are mixed with the original training set to train new models.  Finally, 892 pseudo label data are selected and mixed with the training set to train the final models.  The regression model is jointly trained with the classification models.  The models that performed well in cross-validation are selected and averaged by the weighted fusion based on the confidence. 

}

\begin{table}[!htb]
\centering
\begin{tabular}{cccc}
\hline \textbf{Models} & \textbf{AT} & \textbf{LC}& \textbf{AT + LC} \\ \hline
BERT & 0.9459& 0.9490& \textbf{0.9534}\\
RoBERTa & 0.9480& 0.9482& \textbf{0.9569}\\
XLNet & 0.9462& \textbf{0.9487}& 0.9470\\
ERNIE & 0.9491& 0.9499& \textbf{0.9512}\\
\hline
\end{tabular}
\caption{\label{tab:AT+LC} The performance (accuracy) of task 1a with different training strategies.  
}
\end{table}


\begin{table}[!htb]
\begin{tabular}{c@{~~}c@{~~}c@{~~}c}
\hline \textbf{Models} & \textbf{1a (Acc.)} & \textbf{1a (F1)} & \textbf{1b (RMSE)} \\ \hline
ST & 0.9569& 0.9470& 0.6059\\
MT & 0.9577& 0.9480& 0.5823\\
MT+WL & \textbf{0.9637}& \textbf{0.9550}& \textbf{0.5701}\\
\hline
\end{tabular}
\centering
\caption{\label{tab:strategies} Comparison of different strategies.  ST: single task.  MT: multi-task. WL: weigh loss.}
\end{table}
\subsection{Results}
{
In order to prove the effectiveness of adversarial training (AT) and loss correction (LC), we verify task 1a on four pre-training models.  AT denotes the models through adversarial training by adding perturbations in the embedding layer.  LC denotes the strategy to make correction on the classification cross entropy to interfere with the influence of noise on the model.  AT+LC means to apply both strategies in the training.  Results reported in Table~\ref{tab:strategies} show that by employing individual  strategy, the models can attain good performance on task 1a while employing both strategies can gain better accuracy in BERT, RoBERTa, and ERNIE.

Moreover, we verify the effectiveness of the interactive multi-task training strategy on RoBERTa.  MT+WL denotes that the weighted hyperparameters in the loss function are adjusted based on uncertainty, determined by the learned $\sigma_1$'s and $\sigma_2$, during interactive multi-task training to scale the output the loss function of each task in a similar range.  Results reported Table~\ref{tab:strategies} show that the multi-task joint training mechanism can reduce the RMSE of the regression task (i.e., 1b) significantly while adjusting the loss weight can further decrease the error.

Finally, we attain the F1 score of 0.9570 and the accuracy of 0.9653 on task 1a, respectively.  The RMSE on task 1b is 0.5572.  The RMSE on task 2 is 0.446.
}
\section{Conclusion and Future Work}
This paper presents our system for SemEval-2021 task 7.  Several techniques, such as interactive multi-task joint training, adversarial training, and loss correction, are applied to tackle the task.  More specifically, the perturbation is first added to the input embedding layer and the predicted labels are also added with the real labels to reduce the loss of the noise point data.  Next, the output of task 1a by the Softmax is concatenated with the input of the task 1b to perform joint training on both  tasks.  {Meanwhile, the uncertainty weighting scheme on the loss allows the simple task to have a higher weight.  Finally, multiple models are ensembled to yield the final prediction results. }  Our system attains the first place in the competition. 

In the future, we can explore and verify three other effective strategies.  The first strategy is the task-adaptive funetuning on the pre-trained language models.  Relevant sentences can be  continuously fed into the pre-trained language models to improve the model performance.   The second strategy is to build a graph neural network (GNN) model to exploit all vocabulary for text classification.  Because BERT is relatively limited to capture the global information from a larger language vocabulary, it is promising to facilitate the GNN, which captures the global information, with the in-depth interaction of BERT's middle layers, which embed sufficient local information.  
We will further investigate discourse structures~\cite{lei2017swim, lei2018linguistic} for humor detection. Because, both BERT and GNN models information from word relations, it is necessary to involve the study of discourse structures, which describe how two sentences are logically connected to one another.
By such novel design, we can attain better representations and improve the classification performance. 

\bibliographystyle{acl_natbib}
\bibliography{anthology,acl2021}

\begin{thebibliography}{28}
\expandafter\ifx\csname natexlab\endcsname\relax\def\natexlab#1{#1}\fi

\bibitem[{Castro et~al.(2016)Castro, Cubero, Garat, and Moncecchi}]{5}
Santiago Castro, Mat{\'i}as Cubero, Diego Garat, and Guillermo Moncecchi. 2016.
\newblock \href {https://arxiv.org/abs/1703.09527} {Is this a joke? detecting
  humor in spanish tweets}.
\newblock In \emph{Advances in Artificial Intelligence - IBERAMIA 2016}, pages
  139--150, Cham. Springer International Publishing.

\bibitem[{Chen and Soo(2018)}]{7}
Peng-Yu Chen and Von-Wun Soo. 2018.
\newblock \href {https://doi.org/10.18653/v1/N18-2018} {Humor recognition using
  deep learning}.
\newblock In \emph{Proceedings of the 2018 Conference of the North {A}merican
  Chapter of the Association for Computational Linguistics: Human Language
  Technologies, Volume 2 (Short Papers)}, pages 113--117, New Orleans,
  Louisiana. Association for Computational Linguistics.

\bibitem[{Fan et~al.(2020)Fan, Lin, Yang, Diao, Shen, Chu, and Zou}]{11}
Xiaochao Fan, Hongfei Lin, Liang Yang, Yufeng Diao, Chen Shen, Yonghe Chu, and
  Yanbo Zou. 2020.
\newblock \href {https://doi.org/https://doi.org/10.1016/j.neucom.2020.02.030}
  {Humor detection via an internal and external neural network}.
\newblock \emph{Neurocomputing}, 394:105--111.

\bibitem[{Jiao et~al.(2019)Jiao, Yang, King, and
  Lyu}]{DBLP:conf/naacl/JiaoYKL19}
Wenxiang Jiao, Haiqin Yang, Irwin King, and Michael~R. Lyu. 2019.
\newblock \href {https://doi.org/10.18653/v1/n19-1037} {Higru: Hierarchical
  gated recurrent units for utterance-level emotion recognition}.
\newblock In \emph{{NAACL-HLT}}, pages 397--406. Association for Computational
  Linguistics.

\bibitem[{Kendall et~al.(2017)Kendall, Gal, and Cipolla}]{16}
Alex Kendall, Yarin Gal, and Roberto Cipolla. 2017.
\newblock \href {http://arxiv.org/abs/1705.07115} {Multi-task learning using
  uncertainty to weigh losses for scene geometry and semantics}.
\newblock \emph{CoRR}, abs/1705.07115.

\bibitem[{Khodak et~al.(2018)Khodak, Saunshi, and Vodrahalli}]{18}
Mikhail Khodak, Nikunj Saunshi, and Kiran Vodrahalli. 2018.
\newblock \href {https://www.aclweb.org/anthology/L18-1102} {A large
  self-annotated corpus for sarcasm}.
\newblock In \emph{{LREC}}, Miyazaki, Japan. European Language Resources
  Association (ELRA).

\bibitem[{Kiddon and Brun(2011)}]{4}
Chlo{\'e} Kiddon and Yuriy Brun. 2011.
\newblock \href {https://www.aclweb.org/anthology/P11-2016} {That{'}s what she
  said: Double entendre identification}.
\newblock In \emph{Proceedings of the 49th Annual Meeting of the Association
  for Computational Linguistics: Human Language Technologies}, pages 89--94,
  Portland, Oregon, USA. Association for Computational Linguistics.

\bibitem[{Lefcourt and Martin(1986)}]{19}
Herbert~M. Lefcourt and Rod~A. Martin. 1986.
\newblock \emph{Humor and Life Stress}.
\newblock Springer New York.

\bibitem[{Lei et~al.(2020)Lei, He, de~Rijke, and Chua}]{lei2020conversational}
Wenqiang Lei, Xiangnan He, Maarten de~Rijke, and Tat-Seng Chua. 2020.
\newblock Conversational recommendation: Formulation, methods, and evaluation.
\newblock In \emph{Proceedings of the 43rd International ACM SIGIR Conference
  on Research and Development in Information Retrieval}, pages 2425--2428.

\bibitem[{Lei et~al.(2017)Lei, Wang, Liu, Ilievski, He, and Kan}]{lei2017swim}
Wenqiang Lei, Xuancong Wang, Meichun Liu, Ilija Ilievski, Xiangnan He, and
  Min-Yen Kan. 2017.
\newblock Swim: A simple word interaction model for implicit discourse relation
  recognition.
\newblock In \emph{IJCAI}, pages 4026--4032.

\bibitem[{Lei et~al.(2018)Lei, Xiang, Wang, Zhong, Liu, and
  Kan}]{lei2018linguistic}
Wenqiang Lei, Yuanxin Xiang, Yuwei Wang, Qian Zhong, Meichun Liu, and Min-Yen
  Kan. 2018.
\newblock Linguistic properties matter for implicit discourse relation
  recognition: Combining semantic interaction, topic continuity and
  attribution.
\newblock In \emph{Proceedings of the AAAI Conference on Artificial
  Intelligence}.

\bibitem[{Liang et~al.(2020)Liang, Lei, Chan, Yang, Sun, and
  Chua}]{liang2020pirhdy}
Hongru Liang, Wenqiang Lei, Paul~Yaozhu Chan, Zhenglu Yang, Maosong Sun, and
  Tat-Seng Chua. 2020.
\newblock Pirhdy: Learning pitch-, rhythm-, and dynamics-aware embeddings for
  symbolic music.
\newblock In \emph{Proceedings of the 28th ACM International Conference on
  Multimedia}, pages 574--582.

\bibitem[{Liu et~al.(2018)Liu, Zhang, and Song}]{8}
Lizhen Liu, Donghai Zhang, and Wei Song. 2018.
\newblock \href {https://doi.org/10.18653/v1/P18-2093} {Modeling sentiment
  association in discourse for humor recognition}.
\newblock In \emph{Proceedings of the 56th Annual Meeting of the Association
  for Computational Linguistics (Volume 2: Short Papers)}, pages 586--591,
  Melbourne, Australia. Association for Computational Linguistics.

\bibitem[{Liu et~al.(2020)Liu, Cheng, He, Chen, Wang, Poon, and Gao}]{14}
Xiaodong Liu, Hao Cheng, Pengcheng He, Weizhu Chen, Yu~Wang, Hoifung Poon, and
  Jianfeng Gao. 2020.
\newblock \href {http://arxiv.org/abs/2004.08994} {Adversarial training for
  large neural language models}.

\bibitem[{Ma et~al.(2020)Ma, Xie, Jin, Lianxin, Yang, and Shen}]{10}
Jian Ma, ShuYi Xie, Meizhi Jin, Jiang Lianxin, Mo~Yang, and Jianping Shen.
  2020.
\newblock \href {https://www.aclweb.org/anthology/2020.semeval-1.142}
  {{XSYSIGMA} at {S}em{E}val-2020 task 7: Method for predicting headlines{'}
  humor based on auxiliary sentences with {EI}-{BERT}}.
\newblock In \emph{Proceedings of the Fourteenth Workshop on Semantic
  Evaluation}, pages 1077--1084, Barcelona (online). International Committee
  for Computational Linguistics.

\bibitem[{Meaney et~al.(2021)Meaney, Wilson, Chiruzzo, and
  Magdy}]{meaney2021hahackathon}
J.A. Meaney, Steven~R. Wilson, Luis Chiruzzo, and Walid Magdy. 2021.
\newblock Semeval 2021 task7, hahackathon, detecting and rating humor and
  offense.
\newblock In \emph{Proceedings of the 59th Annual Meeting of the Association
  for Computational Linguistics and the 11th International Joint Conference on
  Natural Language Processing}.

\bibitem[{Mihalcea and Strapparava(2005)}]{2}
Rada Mihalcea and Carlo Strapparava. 2005.
\newblock \href {https://www.aclweb.org/anthology/H05-1067} {Making computers
  laugh: Investigations in automatic humor recognition}.
\newblock In \emph{Proceedings of Human Language Technology Conference and
  Conference on Empirical Methods in Natural Language Processing}, pages
  531--538, Vancouver, British Columbia, Canada. Association for Computational
  Linguistics.

\bibitem[{{Mikolov} et~al.(2011){Mikolov}, {Kombrink}, {Burget}, {Černocký},
  and {Khudanpur}}]{3}
T.~{Mikolov}, S.~{Kombrink}, L.~{Burget}, J.~{Černocký}, and S.~{Khudanpur}.
  2011.
\newblock \href {https://doi.org/10.1109/ICASSP.2011.5947611} {Extensions of
  recurrent neural network language model}.
\newblock In \emph{ICASSP}, pages 5528--5531.

\bibitem[{Miyato et~al.(2017)Miyato, Dai, and Goodfellow}]{13}
Takeru Miyato, Andrew~M. Dai, and Ian Goodfellow. 2017.
\newblock \href {https://arxiv.org/abs/1605.07725} {Adversarial training
  methods for semi-supervised text classification}.
\newblock \emph{Machine Learning}.

\bibitem[{Nijholt et~al.(2003)Nijholt, Stock, Dix, and Morkes}]{20}
Anton Nijholt, Oliviero Stock, Alan Dix, and John Morkes. 2003.
\newblock \href {https://doi.org/10.1145/765891.766143} {Humor modeling in the
  interface}.
\newblock \emph{Conference on Human Factors in Computing Systems -
  Proceedings}.

\bibitem[{Pan et~al.(2019)Pan, Lei, Chua, and Kan}]{pan2019recent}
Liangming Pan, Wenqiang Lei, Tat-Seng Chua, and Min-Yen Kan. 2019.
\newblock Recent advances in neural question generation.
\newblock \emph{arXiv preprint arXiv:1905.08949}.

\bibitem[{Reed et~al.(2015)Reed, Lee, Anguelov, Szegedy, Erhan, and
  Rabinovich}]{17}
Scott Reed, Honglak Lee, Dragomir Anguelov, Christian Szegedy, Dumitru Erhan,
  and Andrew Rabinovich. 2015.
\newblock \href {https://arxiv.org/abs/1412.6596v1} {Training deep neural
  networks on noisy labels with bootstrapping}.
\newblock \emph{Computer Vision and Pattern Recognition}, abs/1705.07115.

\bibitem[{Szegedy et~al.(2014)Szegedy, Zaremba, Sutskever, Bruna, Erhan,
  Goodfellow, and Fergus}]{12}
Christian Szegedy, Wojciech Zaremba, Ilya Sutskever, Joan Bruna, Dumitru Erhan,
  Ian Goodfellow, and Rob Fergus. 2014.
\newblock \href {https://arxiv.org/abs/1312.6199} {Intriguing properties of
  neural networks}.
\newblock \emph{Computer Vision and Pattern Recognition}.

\bibitem[{Taylor and Mazlack(2004)}]{1}
Julia~M. Taylor and Lawrence~J. Mazlack. 2004.
\newblock \href {https://escholarship.org/uc/item/0v54b9jk} {Computationally
  recognizing wordplay in jokes}.
\newblock \emph{Proceedings of the Annual Meeting of the Cognitive Science
  Society}, 26.

\bibitem[{Weller and Seppi(2019)}]{9}
Orion Weller and Kevin~D. Seppi. 2019.
\newblock \href {http://arxiv.org/abs/1909.00252} {Humor detection: {A}
  transformer gets the last laugh}.
\newblock \emph{CoRR}, abs/1909.00252.

\bibitem[{Xia and Ding(2019)}]{15}
Rui Xia and Zixiang Ding. 2019.
\newblock \href {https://doi.org/10.18653/v1/P19-1096} {Emotion-cause pair
  extraction: A new task to emotion analysis in texts}.
\newblock In \emph{Proceedings of the 57th Annual Meeting of the Association
  for Computational Linguistics}, pages 1003--1012, Florence, Italy.
  Association for Computational Linguistics.

\bibitem[{Yan and Pedersen(2017)}]{6}
Xinru Yan and Ted Pedersen. 2017.
\newblock \href {http://arxiv.org/abs/1704.08390} {Duluth at semeval-2017 task
  6: Language models in humor detection}.
\newblock \emph{CoRR}, abs/1704.08390.

\bibitem[{Zhang et~al.(2020)Zhang, Zhang, Wang, Liang, Lei, Sun, Jatowt, and
  Yang}]{zhang2020generalized}
Yao Zhang, Xu~Zhang, Jun Wang, Hongru Liang, Wenqiang Lei, Zhe Sun, Adam
  Jatowt, and Zhenglu Yang. 2020.
\newblock Generalized relation learning with semantic correlation awareness for
  link prediction.
\newblock \emph{arXiv preprint arXiv:2012.11957}.

\end{thebibliography}


\end{document}